\documentclass{article} 
\usepackage{iclr2023_workshop,times}

\usepackage{amsmath,amsfonts,bm}

\def\eqref#1{equation~\ref{#1}}

\def\1{\bm{1}}

\DeclareMathAlphabet{\mathsfit}{\encodingdefault}{\sfdefault}{m}{sl}
\SetMathAlphabet{\mathsfit}{bold}{\encodingdefault}{\sfdefault}{bx}{n}

\newcommand{\R}{\mathbb{R}}

\usepackage{hyperref}
\usepackage{url}
\usepackage{svg}
\usepackage{booktabs}
\usepackage{makecell}

\usepackage{tikz}
\usepackage{mathdots}
\usepackage{yhmath}
\usepackage{cancel}
\usepackage{color}
\usepackage{siunitx}
\usepackage{array}
\usepackage{multirow}
\usepackage{amssymb}
\usepackage{tabularx}
\usepackage{extarrows}
\usepackage{booktabs}
\usetikzlibrary{fadings}
\usetikzlibrary{patterns}
\usetikzlibrary{shadows.blur}
\usetikzlibrary{shapes}
\usepackage{standalone}

\newcommand{\fakesection}[1]{%
  \par\refstepcounter{section}
  \sectionmark{#1}
  \addcontentsline{toc}{section}{\protect\numberline{\thesection}#1}
}

\usepackage{siunitx}
\title{Multi-Scale Message Passing Neural PDE Solvers}

\author{Léonard Equer \\
ETH Z\"urich
\And
T. Konstantin Rusch \\
ETH Z\"urich and UC Berkeley\\
\texttt{trusch@ethz.ch}
\And
Siddhartha Mishra\\
ETH Z\"urich
}

\iclrfinalcopy 
\begin{document}

\maketitle

\begin{abstract}
We propose a novel \emph{multi-scale message passing neural network algorithm} for learning the solutions of time-dependent PDEs. Our algorithm possesses both temporal and spatial multi-scale resolution features by incorporating multi-scale sequence models and graph gating modules in the encoder and processor, respectively. Benchmark numerical experiments are presented to demonstrate that the proposed algorithm outperforms baselines, particularly on a PDE with a range of spatial and temporal scales. 
\end{abstract}

\section{Introduction}
Time-dependent partial differential equations (PDEs) arise as mathematical models of many interesting phenomena in the sciences and engineering that involve the time-evolution of physical quantities of interest \citep{Evansbook}. Solving such PDEs entails computing the so-called \emph{solution operator} that maps the initial conditions (and other inputs such as coefficients, sources etc) to the trajectories of the solution over time. \emph{Classical} numerical methods which combine spatial discretizations such as finite differences, finite elements or spectral methods together with Runge-Kutta or multi-step temporal discretization schemes are widely used to simulate time-dependent PDEs \citep{NAbook}. However, these methods can be prohibitively expensive, particularly in several spatial dimensions and for long-time integration. 

Recently, machine learning based algorithms are increasingly being used for the fast and accurate simulation of time-dependent PDEs. Examples include supervised learning algorithms \citep{ZhuZab1,LMR1}, physics informed neural networks \citep{KAR1,KAR2} and operator learning algorithms such as DeepONets \citep{donet} and Fourier Neural Operators \citep{FNO}. However, each of these frameworks raises many unaddressed issues in terms of applicability, efficiency and generalization capacity. 

In particular, existing frameworks such as FNOs or CNNs rely on input (and output) data on uniform Cartesian grids whereas in practice for engineering applications, data generated from simulations or observations is only available on unstructured grids. Given this discrepancy, learning frameworks that admit inputs and outputs on general and highly variable grids could be useful as PDE solvers. In this context, models based on graph neural networks are increasingly being considered as attractive frameworks for learning PDEs \citep{MGN1,bengioneural,Bat1} and references therein. 

A very attractive framework in this regard was recently proposed in \citet{messpassnPDE}, where the authors suggested an autoregressive message passing procedure to learn solution operators of time-dependent PDEs. The approach utilized the well-known \emph{encode-process-decode} paradigm \citep{Bat1} by embedding time-dependent inputs into a feature vector on a graph (related to the underlying possibly unstructured grid), processing this feature vector through message passing graph neural networks (GNNs) and mapping this latent representation into a time-update through a decoder. This approach was shown to be competitive vis a vis other models for some representative examples, particularly of PDEs where the solution only has a small range of scales in both space and time. 

However in practice, solutions to a large class of time-dependent PDEs contain structures at a wide range of scales in both space and time. Such \emph{multi-scale} PDEs \citep{MSbook} include the well-known models of fluid dynamics, wave propagation and reaction-diffusion mechanisms. Our aim in this paper is to propose a machine learning framework that can accurately learn the solutions of such multiscale time-dependent PDEs. To this end, we base our algorithm on the graph learning based framework of \citet{messpassnPDE} but endow it with \emph{multi-scale} features. In particular, the encoder is supplemented with a very recent multi-scale sequence modeling algorithm called the \emph{long expressive memory} (LEM) \citep{LEM}. Similarly, instead of employing standard GNNs as processors, we modify them by a novel \emph{gating} mechanism, analogous to the one suggested recently in \citet{g2}. We demonstrate through numerical experiments that these multi-scale augmentations not only improve performance on standard single-scale benchmarks but also significantly outperform competing models on a multi-scale time-dependent PDE. These promising results pave the way for the design of a robust and accurate autoregressive message passing framework for learning time-dependent PDEs. 
\section{The Method.}
\paragraph{Setting.} We consider the following abstract form of a time-dependent PDE,
\begin{equation}
\begin{aligned}
\partial_t u = \mathcal{N}_{\eta}(u), \quad  \text { in } \Omega \times]0, T[ \\
u = u_0(x) \quad \text { in } \bar{\Omega} \times\{0\}
\end{aligned}
\label{pbstatement}
\vspace*{-0.15cm}
\end{equation}
Here, $\Omega \in \R^d$ is a bounded open set and $T >0$. The differential operator $\mathcal{N}_\eta:\mathcal{H} \mapsto \mathcal{\bar{H}}$, maps between two Hilbert spaces $\mathcal{H},\mathcal{\bar{H}}$ and $\eta$ models a coefficient, which for simplicity we assume to be finite-dimensional $\eta \in \R^{d_\eta}$. Finally the initial data is $u_0 \in \mathcal{H}$ and the PDE \eqref{pbstatement} is augmented with suitable boundary conditions. Our objective is to learn the solution operator
\begin{equation}
\label{eq:solop}
\mathcal{S}^\eta_t: \mathcal{H} \mapsto \mathcal{H}, \quad u(.,t) = \mathcal{S}^\eta_t u_0.
\vspace*{-0.05cm}
\end{equation}
To this end, we will proceed by learning the \emph{autoregressive} mapping
\begin{equation}
u(.,t+\Delta t)=\mathcal{A}_{\eta}^{\Delta t}(u(.,t)) 
\vspace*{-0.1cm}
\end{equation}
In other words, the above operator maps the solution $u(.,t)$ of \eqref{pbstatement} at current time $t$ to the solution at a later time $t+\Delta t$. By iteratively applying $\mathcal{A}_{\eta}^{\Delta t}$, we can extend the solution over the entire time period. We drop the $\Delta t$-dependence below for notational convenience.

In order to learn this mapping, we define a grid $\{x_n \in \Omega \}$ such that we obtain a finite dimensional approximation of $u(.,t) \in \mathcal{H}$ as $\mathbf{u}(t) = [u(t,x_1), \ldots , u(t,x_N)]^\top \in \mathbb{R}^{n_x}$. We can now approximate the infinite dimensional operator $\mathcal{A}_{\eta}$ by a finite dimensional operator $\mathcal{A}_{\eta,\theta}$ parameterized by $\theta \in \Theta \subset \mathbb{R}^D$ as
\begin{equation}
\label{eq:autoreg}
\vspace*{-0.2cm}
\mathbf{u}(t + \Delta t)= \mathcal{A}_{\eta,\theta}(\mathbf{u}(t))
\end{equation}

Next, we will describe the key ingredients of our multi-scale message-passing paradigm to learn the autoregressive map \ref{eq:autoreg}. 
\paragraph{The message passing framework of \citet{messpassnPDE}.}
We start by a brief description of the message passing framework of \citet{messpassnPDE}, which we will augment with multi-scale features later. 

As mentioned before, this framework follows a \emph{encode-process-decode} paradigm. The input to the {\bf encoder} is the vector of K-lagged \footnote{A K-lagged solution $\mathbf{u}^{i-K:i}$  is defined as the  following set of vectors $\{\mathbf{u}(t_{i - K}), \mathbf{u}(t_{i - K + 1}), ... , \mathbf{u}(t_{i-1})\}$} solutions $\mathbf{u}^{k-K:k}$, containing the (recent) history of the solution trajectory. This vector, at each grid point, is then \emph{embedded} into a high-dimensional feature vector $\mathbf{X}^0_i$ at each node $i$ of a \emph{Graph} $\mathcal{G}$, which in turn, is defined in terms of the underlying grid points forming nodes $\{i\}_i$ and sets of nearest neighbors $\{\mathcal{N}(i)\}_i$ constituting edges (see {\bf SM} \ref{network_architecture} for details of this computational graph). The encoder mapping of \cite{messpassnPDE} is a shallow neural network. Next, the feature vector $\mathbf{X}^0_i$ is augmented with relative positions $x_i -x_j$, the equation parameters $\eta$ as well as the solution differences $\mathbf{u}^{k-K:k}_i - \mathbf{u}^{k-K:k}_j$ and is processed through a multi-hidden layer \emph{message passing neural network} (MPNN) \citep{mpnn}, with the relative positions, parameters and solution finite differences being fed as inputs to each hidden layer. The output of the last hidden layer $\{{\bf X}^L_i\}_i$ is then transformed into the update $\mathbf{u}^{k:k+K}$ that provides future trajectories of the solution. The {\bf decoder} is a one-dimensional convolutional neural network. 
\vspace*{-0.05cm}
Summarizing the message passing framework of \citet{messpassnPDE} yields the following mapping, 
\begin{equation}
\label{eq:model}
\mathbf{u}^{k:k+K} = \mathcal{A}_{\eta,\theta}(\mathbf{u}^{k-K:k}, \mathcal{G})
\end{equation}
for updating the solution trajectories of the time-dependent PDE \eqref{pbstatement}. 

\paragraph{Resolving multiple time scales with long-expressive memory (LEM).}
Long expressive memory (LEM) was proposed recently in \citet{LEM} as a sequence model that can i) learn long-term dependencies in sequential data as it solves the exploding and vanishing gradient problem and ii) it can efficiently process multiple scales in the data. It is this latter feature that we seek to exploit in our context. LEM is based on the structure preserving implicit-explicit discretization of an ODE system such that the recurrent update rule becomes
\begin{equation}
\label{eq:lem}
\begin{aligned}
\Delta \mathbf{t}_n & =\Delta t \hat{\sigma}\left(\mathbf{W}_1 \mathbf{y}_{n-1}+\mathbf{V}_1 \mathbf{u}_n+\mathbf{b}_1\right) \\
\overline{\Delta \mathbf{t}_n} & =\Delta t \hat{\sigma}\left(\mathbf{W}_2 \mathbf{y}_{n-1}+\mathbf{V}_2 \mathbf{u}_n+\mathbf{b}_2\right) \\
\mathbf{z}_n & =\left(1-\boldsymbol{\Delta} \mathbf{t}_n\right) \odot \mathbf{z}_{n-1}+\Delta \mathbf{t}_n \odot \sigma\left(\mathbf{W}_z \mathbf{y}_{n-1}+\mathbf{V}_z \mathbf{u}_n+\mathbf{b}_z\right) \\
\mathbf{y}_n & =\left(1-\overline{\boldsymbol{\Delta}}_n\right) \odot \mathbf{y}_{n-1}+\overline{\boldsymbol{\Delta}}_n \odot \sigma\left(\mathbf{W}_y \mathbf{z}_n+\mathbf{V}_y \mathbf{u}_n+\mathbf{b}_y\right)
\end{aligned}
\end{equation}
where $\hat{\sigma}$ is a sigmoid activation function, $\sigma(u) = \text{tanh}(u)$, $\mathbf{W}_{1,2,y,z}$ and $\mathbf{V}_{1,2,y,z}$ are weight matrices and $\mathbf{b}_{1,2,y,z}$ bias vectors. The discretized system evolves the \emph{hidden states} by the update formula $\mathbf{z_n}, \mathbf{y_n} = \mathbf{f} ( \mathbf{z_{n-1}}, \mathbf{y_{n-1}}, \mathbf{u_n}, \Delta t )$. We will apply LEM in the encoder step of our proposed architecture to embed our input vector into an expressive high dimensional feature vector.
\paragraph{Resolving multiple spatial scales with Graph Gating.}
Following the recent paper \citet{g2}, we can endow a GNN with the explicit ability to resolve multiple spatial scales by adding a \emph{gating} mechanism. To this end, let $\mathbf{X} \in \mathbb{R}^{N \times N_{\text{hid}}}$ be a feature matrix, $\mathcal{G}$ the graph representation and ${F}_{\hat{\theta}}$, ${F}_{\theta}$ two MPNN with different weights. We can now represent the MPNN updates as
\begin{equation}
\label{eq:gate}
    \mathbf{X}^n = (\mathbf{1} - \hat{\sigma}\left(\mathbf{F}_{\hat{\theta}}(\mathbf{X}^{n-1},\mathcal{G})\right))\odot\mathbf{X}^{n-1} + \hat{\sigma}\left(\mathbf{F}_{\hat{\theta}}(\mathbf{X}^{n-1},\mathcal{G})\right) \odot \sigma(\mathbf{F}_{\theta}(\mathbf{X}^{n-1},\mathcal{G}))
\end{equation} 
Note that in this case we create the two graph neural networks with the same architecture, ${F}_{\hat{\theta}}(\mathbf{X},\mathcal{G})$ learns the scales in the data to act as gating switches when its output are normalized between $0$ and $1$ while ${F}_{\theta}(\mathbf{X},\mathcal{G})$ updates the feature vector according to the message passing rule. 

\begin{figure}[h]
\includegraphics[width=1.0\linewidth]{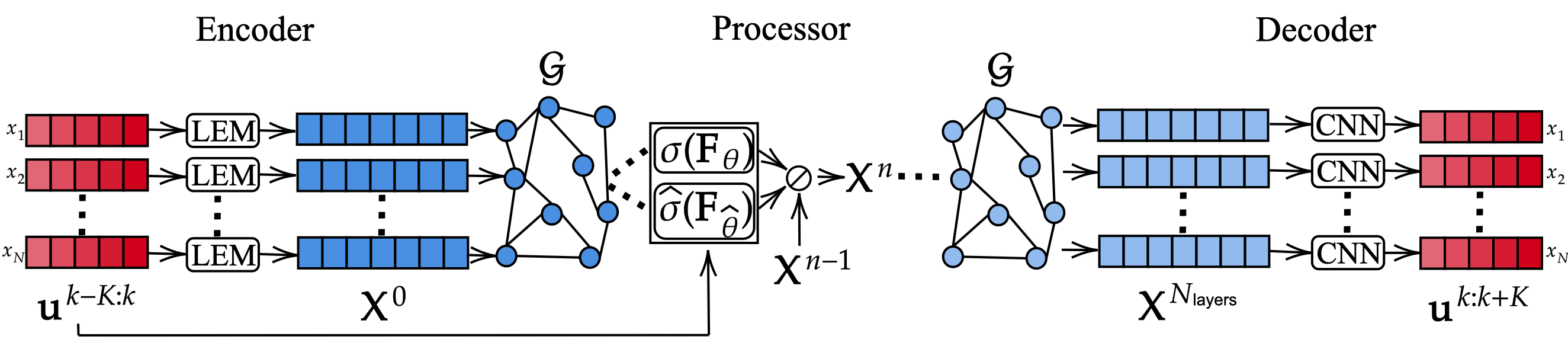}
\centering
\caption{Multi-scale message passing architecture, the encoder applies a LEM on each node input to generate the node embeddings, the processor then performs $N_{\text{layers}}$ of message passing with gating ($\oslash (a,b,c) = (1-b)\odot a + b\odot c$). The node features are then passed through a 1D convolutional layer to project back to the input dimension.}
\label{arch}
\end{figure}

\paragraph{The multi-scale message passing algorithm.} We combine the above three ingredients to form the multi-scale message passing neural PDE solver, which is summarized in Figure \ref{arch}. We make the following changes to the message passing architecture of \cite{messpassnPDE}: i) in the encoder step, an additional LEM layer is introduced that processes the input vector $\mathbf{u}^{k-K:k}$, at each node, with a LEM recurrent neural network that processes this input to resolve multiple time scales and ii) in the processor step, we augment the standard message passing neural network with a gating neural network as in \eqref{eq:gate} to endow the architecture to resolve multiple spatial scales (see \textbf{SM} \ref{network_architecture}). 
\section{Results}
We test the proposed multi-scale message passing procedure on 3 benchmark experiments below. As a baseline, we will use the message passing algorithm of \citet{messpassnPDE} that we abbreviate as {\bf MP-PDE}. Our multi-scale variant, which uses LEM as a component of the encoder and a gating GNN is abbreviated as {\bf MSMP-PDE}. As additional baselines, we ablate different parts of the proposed algorithm namely, remove the LEM component to obtain {\bf Gated}, remove the gating to obtain {\bf LEM}, replace LEM with LSTM in {\bf LEM} and in {\bf MSMP-PDE} to obtain {\bf LSTM } and {\bf LSTMGated}, respectively. 

\vspace{-1.em}
\begin{table}[h]
\begin{center}
\caption{Test $L^2$ (mean $\pm$ std) relative errors for competing models in all the experiments. The two-best performing models are highlighted in bold.}
\vspace{0.5em}
  \begin{tabular}{l c c c}
    \toprule
     &
    \bfseries \textbf{E1} &
      \bfseries \textbf{E2} &
      \bfseries \textbf{MS-wave} \\
   
    \midrule
    \textbf{MP-PDE} & $0.472 \% \pm 0.037 \% $ & $7.309 \% \pm 0.484 \% $ &  $20.39 \% \pm 1.64 \% $  \\
    \textbf{LSTM} & $0.587 \% \pm 0.137 \% $ & $6.788 \% \pm 0.176 \% $ &  $15.84 \% \pm 1.27 \% $  \\
    \textbf{LEM} & $0.455 \% \pm 0.026 \% $ & $6.866 \% \pm 0.235 \% $ &  $17.08 \% \pm 1.45 \% $  \\
    \textbf{Gated} & $0.364 \% \pm 0.051 \% $ & $6.469 \% \pm 0.259 \% $  &  ${\bf 11.9 \% \pm 1.22 \% }$  \\
    \textbf{LSTMGated} & ${\bf 0.326 \% \pm 0.026 \%} $ & ${\bf 6.207 \% \pm 0.345 \% }$ &  $12.49 \% \pm 2.79 \% $  \\
    \textbf{MSMP-PDE} & ${\bf 0.323 \% \pm 0.034 \% }$ & ${\bf 6.302 \% \pm 0.373 \%} $ &  ${\bf 10.36 \% \pm 0.99 \%} $  \\
    \bottomrule
  \end{tabular}
\label{tab:results}
\end{center}
\end{table}
\vspace{-0.5em}
For both the first and second numerical experiments, we will consider Burgers' equation (see {\bf SM} \ref{burger_experiment}) on the computational domain $[0,L]$. In the first experiment, that we abbreviate as {\bf E1}, we follow \citet{messpassnPDE} to consider the inviscid Burgers' equation with the sinusoidal initial data given in {\bf SM} \eqref{burgsetup}. The training (and test) data are generated for different initial conditions with a finite volume scheme and we train all the models on 2048 samples and test them on 128 samples. The test errors are presented in Table \ref{tab:results}. We observe from this table that all the models yield very low relative errors. This is not unexpected as the solution in this case (see {\bf SM} Figure \ref{fig:figureE1} for an illustration) has rather simple dynamic evolution. Nevertheless, {\bf MSMP-PDE} still outperforms the baseline {\bf MP-PDE} to some extent. Ablating different components shows that the gating mechanism is of greater importance in this case than resolving temporal multi-scale behavior. 

In the second experiment, labelled as {\bf E2}, we again follow \citet{messpassnPDE} to consider the forced viscous version of Burgers' equation {\bf SM} \eqref{eq:burg} with the same sinusoidal initial conditions as in {\bf E1}. In this case, training and test samples are generated by varying both the initial conditions as well as the time dependent source terms, leading to significantly more complicated dynamic behavior as compared to {\bf E1} (see {\bf SM} Figure \ref{fig:figureE2} for an illustration). Consequently, the test errors for each of the models, reported in Table \ref{tab:results} are much higher. Again, {\bf MSMP-PDE} outperforms the baseline {\bf MP-PDE} significantly, with the test error being reduced by almost $14\%$. Ablating models shows that adding both gating and temporal multi-scale resolution helps reduce the error, with gating playing a bigger role. 

For the final experiment that we label as {\bf MS-wave}, we consider a one-dimensional wave equation, written as a linear hyperbolic system ({\bf SM} \eqref{twospeed}), that was proposed in \citet{HilSM1} to test numerical methods for multi-scale problems. The initial conditions, given in the form of sinusoidal data ({\bf SM} \eqref{burgsetup}) evolve in the form of waves that propagate at different speeds. The training (and test) samples are generated from different initial conditions as well as wave speeds (see details in {\bf SM} \ref{twospeed_exp}). The test errors with the competing models is reported in Table \ref{tab:results} and we observe from this table that the errors are quite high for this problem given the fact that the models have to learn dynamics at different temporal and spatial scales. For instance, the baseline {\bf MP-PDE} algorithm of \citet{messpassnPDE} yields errors greater than $20\%$. On the other hand, adding both temporal and spatial multi-scale resolution features in our proposed {\bf MSMP-PDE} model leads to large (factor of $2$) decrease in this error. Ablating components shows that both gating and a LEM encoder seem necessary to obtain the best results. 

Thus, with these numerical experiments, we have demonstrated that adding temporal and spatial multiscale resolution capabilities to a message passing neural PDE solver significantly enhances its abilities in approximating PDEs, particularly those with multiple scales, accurately. This study paves the way for the further development of operator learning models that can deal with multi-scale data on both structured and unstructured grids.

\newpage
\bibliography{iclr2023_workshop.bib}
\bibliographystyle{iclr2023_workshop.bst}

\newpage

\appendix
\fakesection{Appendix}
\begin{center}
{\bf Supplementary Material for:}\\
Multi-scale message passing neural PDE solvers.
\end{center}
\subsection{Experiments}
\subsubsection{Burger's equation (\textbf{E1}/\textbf{E2})}\label{burger_experiment}
As a first benchmark, we consider Burger's equation with variable viscosity and forcing, i.e., 
\begin{equation} \label{eq:burg}
\left[\partial_t u+\partial_x\left(u^2-\beta \partial_x u\right)\right](t, x)= \alpha f(t, x)
\end{equation}
\begin{equation}
\label{burgsetup}
 u(t, 0)=u(t,L), \quad   u(0, x)= f(0, x), \quad f(t, x)=\sum_{j=1}^J A_j \sin \left(\omega_j t+2 \pi \ell_j x / L+\phi_j\right),
\end{equation}
and perform two experiments:
\begin{itemize}
    \item \textbf{E1} Inviscid Burger's Equation ($\alpha = \beta = 0$, i.e., no forcing and no diffusion).
    \item \textbf{E2} Burger's Equation with variable viscosity and forcing ($\beta \in [0,0.2]$ and $\alpha = 1$).
\end{itemize}

We generate the corresponding training, validation and test sets based on the numerical solver from \cite{messpassnPDE}.
Thereby, a numerical ground truth is generated using a WENO5 scheme for the convection term and a fourth order finite difference for the diffusion term. The space discretization is then integrated with an explicit Runge-Kutta solver (RK4) with adaptive timestepping. The numerical ground truth is generated with $t \in [0,4]$ and $\Omega = [0,16]$ on a $(n_t,n_x) = (250,200)$-grid. We further compute the training data by down-sampling with a 1D convolution operator to obtain a $(n_t,n_x) = (250,100)$-grid.

Initial conditions and forcing term are defined as $f$ in \eqref{burgsetup} with $A_j \sim U([-1/2,1/2])\quad \phi_j\sim U([0,2\pi]), \quad \omega_j \sim U([-0.4,0.4]) \quad \text{and} \quad \l_j \in \{ 1,2,3 \}$\footnote{$U([a,b])$ denotes the uniform distribution with support $[a,b]$.}.

\subsubsection{Linear Advection System (\textbf{MS-wave})} \label{twospeed_exp}
In this experiment, we focus on the following two-speed advection problem in order to test the ability of the models to effectively learn multi-scale properties,
\begin{equation}
    \label{twospeed}
    \partial_t \mathbf{u} + \mathbf{A} \partial_x \mathbf{u} = 0, \quad \mathbf{u}(t,x) = \begin{pmatrix} u^{(1)}(t,x) \\ u^{(2)}(t,x) \end{pmatrix}, \quad \mathbf{u}(t,0) = \mathbf{u}(t,L).
\end{equation}

Since \eqref{twospeed} is a linear hyperbolic system, an analytic solution can be efficiently computed to generate training instances. To this end, in this experiment we consider,
\begin{equation}
    \mathbf{A} = \begin{pmatrix} a+b & b-a \\ b-a & a+b \end{pmatrix}, \quad a,b \in \mathbb{R}.
\end{equation}
Let $\mathbf{A} = \mathbf{R}\mathbf{\Lambda}\mathbf{R^{-1}}$ be its eigendecomposition with, 
\begin{equation}
    \mathbf{\Lambda} = \begin{pmatrix} 2a & 0 \\ 0 & 2b \end{pmatrix}, \quad \mathbf{R} = \begin{pmatrix} -1 & 1 \\ 1 & 1 \end{pmatrix}, \quad \text{and} \quad \mathbf{R}^{-1} = \begin{pmatrix} -1/2 & 1/2 \\ 1/2 & 1/2 \end{pmatrix}.
\end{equation}

Since $\mathbf{\mathbf{\Lambda}}$ is diagonal, \eqref{twospeed} can be written as a system of two uncoupled advection equations,
\begin{equation}
    \partial_t \mathbf{w} + \mathbf{\mathbf{\Lambda}} \partial_x \mathbf{w} = 0, \quad \text{with} \quad \mathbf{w} = \mathbf{R^{-1}}\mathbf{u}.
\end{equation}

Given an initial condition $\mathbf{u}_0(x)$, we can define the initial condition in the eigenbasis,
\begin{equation}
    \mathbf{w}_0(x) = \mathbf{R}^{-1}\mathbf{u}_0(x) = \begin{pmatrix} w^{(1)}_0(x) \\ w^{(2)}_0(x) \end{pmatrix}.
\end{equation}
We can then use the solution of scalar conservation laws by the method of characteristics \citep{leveque_conslaw} to get the solution of the system, 
\begin{equation}
    \mathbf{u}(t,x) = \mathbf{R} \mathbf{w}(t,x) =  \mathbf{R} \begin{pmatrix} w^{(1)}_0(x-2at) \\ w^{(2)}_0(x-2bt) \end{pmatrix}.
\end{equation}

We generate a training, validation and test set using the aforementioned method. The numerical ground truth is generated with  $\Omega = [0,16]$ and $t \in [0,4]$ on a $(n_t,n_x) = (250,200)$-grid and the training data is computed by downsampling to a $(n_t,n_x) = (250,100)$-grid. Initial conditions $u^{(1)}_0$ and $u^{(2)}_0$ are defined as $f$ in \eqref{burgsetup} with $A_j \sim U([-1/2,1/2])\quad \phi_j\sim U([0,2\pi])\quad \text{and} \quad \l_j \in \{ 1,2,3 \}$. The equation parameters are sampled as $a \sim U([0.1,1])$ and $b \sim U([1,10])$ such that two very different scales can be represented.

\subsection{Network Architecture} \label{network_architecture}
In this section, we briefly describe the model implementation.
\paragraph{Encoder.} In the scalar case, the input to the encoder is given by the $(K + \text{dim}(\eta) + 2)$-dimensional vector $[\mathbf{u}_i^{k-K: k}, x_i, t_k, \eta]$. We distinguish between two different types of encoders: (i) a simple two-layer feedforward neural network (with swish activation function) for \textbf{MP-PDE}, (ii) a one-layer LEM (or LSTM) for all other methods considered here, where the $3$-dimensional sequence $\{[u_i^{k-K+l}, x_i,\eta]^\top\}_{l=0}^{K-1}$ of length $K$ gets recurrently processed and only the final hidden state of the RNN gets further propagated through a $2$-layer feedforward neural network (with swish activation) before it gets passed to the message-passing processor.

\paragraph{Processor.} The processor relies on the node feature vectors $\{{\bf X}^n_i\}_i = \mathbf{X}^n$ and graph $\mathcal{G}$ to perform the following message passing operations:
\begin{equation}
\begin{array}{rlr}
\label{mpnn}
\text { edge } j \rightarrow i \text { message: } & \mathbf{m}_{i j}^n=\phi_\theta \left( \left[ {\bf X}_i^{n-1}, {\bf X}_j^{n-1}, \mathbf{u}_i^{k-K: k}-\mathbf{u}_j^{k-K: k}, x_i-x_j, \eta  \right] \right), \\
\text { node } i \text { update: } & \mathbf{F}_{\theta}(\mathbf{X}^{n-1},\mathcal{G})_i=\psi_\theta \left( \left[ {\bf X}_i^{n-1}, \sum_{j \in \mathcal{N}(i)} \mathbf{m}_{i j}^n, \eta \right] \right),
\end{array}
\end{equation}
where $\phi$ and $\psi$ are $2$-layer feedforward neural networks with Swish activation function. For the non-gated models, the node vectors are simply updated as $\mathbf{X}^{n} = \mathbf{F}_{\theta}(\mathbf{X}^{n-1},\mathcal{G})$. In contrast to that, for our proposed gated GNN model we need to construct two MPNNs, i.e., $\mathbf{F}_{\theta}$ and $\mathbf{F}_{\hat{\theta}}$ each defined as in \eqref{mpnn}. We then use the propagation rule in \eqref{eq:gate} to update the node feature vectors.

\paragraph{Decoder.} The decoder is based on a $2$-layer convolutional neural network (CNN) to project the node feature outputs of the final processor layer back to the input-output dimension of the underlying PDE. To this end, we follow \cite{deepmind_auto_cnn,messpassnPDE} and output the difference of the current to the new timestep solution $d_i=\left(d_i^1, d_i^2, \ldots, d_i^K\right)$, i.e., the final node-wise update of our network is, 
\begin{equation}
u_i^{k+\ell}=u_i^k+\left(t_{k+\ell}-t_k\right) d_i^{\ell}, \quad 1 \leq \ell \leq K.
\end{equation}

We summarize the architectures used for 1D scalar PDEs (experiments \textbf{E1}/\textbf{E2}) in table \ref{model_ref}. In practice we fix $N_{\text{hid}} = 128$ with 6 processor hidden layers ($N_{\text{layers}}$).

In order to accommodate 1D systems of $N$ equations (as in experiment \textbf{MS-wave} where $N = 2$) we add a linear layer at the beginning of the decoder that maps the hidden dimension to a vector of size $N \times  K$ such that we can recover the $N$ $K$-lagged output vectors. 

\begin{table}[h!]
\small
\caption{Architecture summary of the tested models.}
\centering
\begin{tabular}{l c c c}
 \toprule
 \bfseries Model & \bfseries Encoder & \bfseries Processor & \bfseries Decoder \\
 \midrule \midrule
 \textbf{MP-PDE} & Linear-Swish-Linear-Swish & 6 $\times$ MPNN & Conv1D-Swish-Conv1D \\
   \textbf{LSTM}  & LSTM-Linear-Swish-Linear-Swish & 6 $\times$ MPNN & Conv1D-Swish-Conv1D \\
 \textbf{LEM}  & LEM-Linear-Swish-Linear-Swish & 6 $\times$ MPNN & Conv1D-Swish-Conv1D \\
  \textbf{Gated}  & Linear-Swish-Linear-Swish & 6 $\times$ Gated MPNN & Conv1D-Swish-Conv1D \\
  \textbf{LSTMGated}  & LSTM-Linear-Swish-Linear-Swish & 6 $\times$ Gated MPNN & Conv1D-Swish-Conv1D \\
  \textbf{MSMP-PDE}  & LEM-Linear-Swish-Linear-Swish & 6 $\times$ Gated MPNN & Conv1D-Swish-Conv1D \\
  
 \bottomrule
 \end{tabular}
 \label{model_ref}
\end{table}

\begin{table}[h!]
\small
\caption{Number of model trainable parameters per experiment.}
\centering
\begin{tabular}{l ccc}
 \toprule
 \textbf{Model} & \textbf{E1} & \textbf{E2} & \textbf{MS-wave} \\
 \midrule\midrule
 \textbf{MP-PDE} & 634'745 & 636'409 & 693'738\\
   \textbf{LSTM}  & 715'769 & 717'817 & 772'842\\
 \textbf{LEM}  & 715'257 & 717'305& 772'330 \\
  \textbf{Gated}  & 1'249'145 & 1'252'345 & 1'330'410 \\
  \textbf{LSTMGated}  & 1'330'169 & 1'333'753 & 1'409'514\\
  \textbf{MSMP-PDE}  & 1'329'657 & 1'333'241 & 1'409'002\\
 \bottomrule
 \end{tabular}
\label{modelparamsnumber}
\end{table}

\subsection{Training and Testing Details}

The training and inference of the models is done on an NVIDIA GeForce RTX 2080 Ti, where the training takes between $24$h and $65$h depending on the model complexity.

\subsubsection{Dataset and Errors}

 In all experiments the domain is given by $\Omega = [0,16]$ and $t \in [0,4]$ with $(n_t,n_x) = (250,100)$ and $K = 25$. 

 For experiments \textbf{E1}/\textbf{E2} we create a training, validation and test set of sizes 2048, 128 and 128 respectively (1024,128,128 for \textbf{MS-wave}). We perform a 5-fold cross validation to get estimates of the standard deviation associated to changing the dataset and resampling the initial weights.  
 The models are trained using the method described in section \ref{training_section}, we use the validation dataset to perform early stopping and report the errors obtained on the test set.
 
 Let $u^i$ and $u^i_{\theta}$ be the ground truth solution and the network prediction for the test sample $i$, we report in table \ref{tab:results} the relative error,
 \begin{equation}
 \text{RE} = \frac{ \frac{1}{N_{\text{sample}}} \sum_{i=1}^{N_{\text{sample}}}\Vert u_\theta^i - u^i\Vert_{L^2(\Omega \times ]0,T])}}{ \frac{1}{N_{\text{sample}}} \sum_{i=1}^{N_{\text{sample}}}\Vert u^i \Vert_{L^2(\Omega \times ]0,T])}},
 \end{equation}
 which in practice is computed by unrolling the full trajectory with the network of interest and by computing the norms in discretized form.

 \subsubsection{Autoregressive Training Details} \label{training_section}
 We follow the training procedure of \citet{messpassnPDE}, i.e., the networks are trained for 20 epochs (with early stopping), we use a batch size of $16$ and a learning rate of \num{1e-4}, which is reduced by a factor of $0.4$ for every $5$ epochs. Moreover, we train the models using the AdamW optimizer with a root mean squared loss. We define the underlying computational graph using a radius around the node of interest such that 3 nearest neighbors on each side are connected to the node. 

The training procedure is illustrated in Figure \ref{fig:train}. We start by grouping each of the temporal sequences of the training set to a set of $K$-lagged solutions. We further randomly choose one of the $K$-lagged solutions as well as the number of unrolling steps to be performed (in this paper we use a maximum unrolling depth of $2$). After unrolling, the loss is computed between the model output and the ground truth solution, where the errors are then backpropagated only through the last model call in order to mitigate the distribution shift problem. \\

\begin{figure}[h]
  \includestandalone[width=\textwidth]{tikz_training}
\caption{Example of three autoregressive training input-output instances with K-lagged solutions, truncated backpropagation and a maximum unrolling depth of 2.}
\label{fig:train}
\end{figure}

\pagebreak

\subsection{Plots}

\begin{figure}[h]

\centering
\includegraphics[width=.5\textwidth]{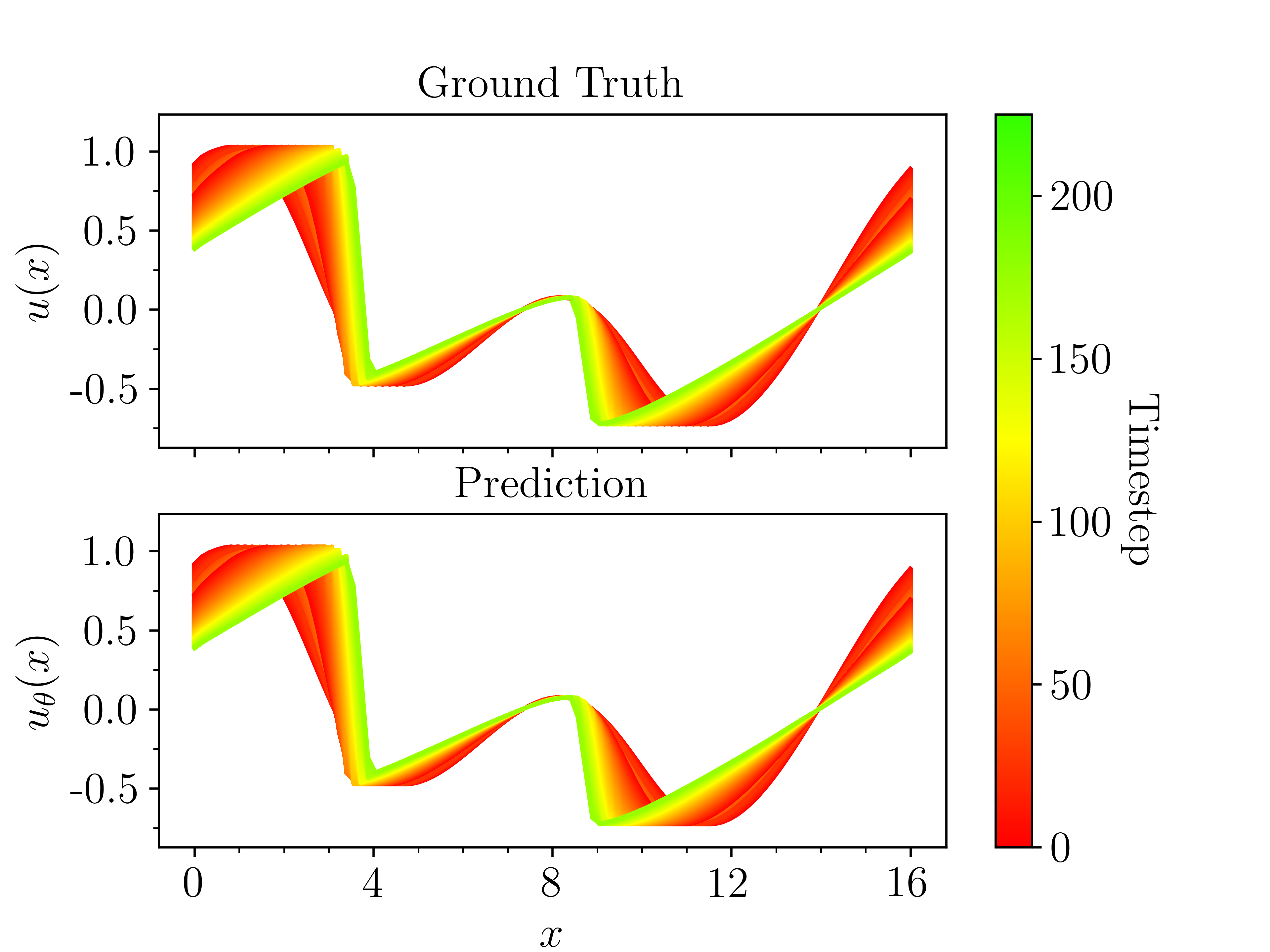}\hfill
\includegraphics[width=.5\textwidth]{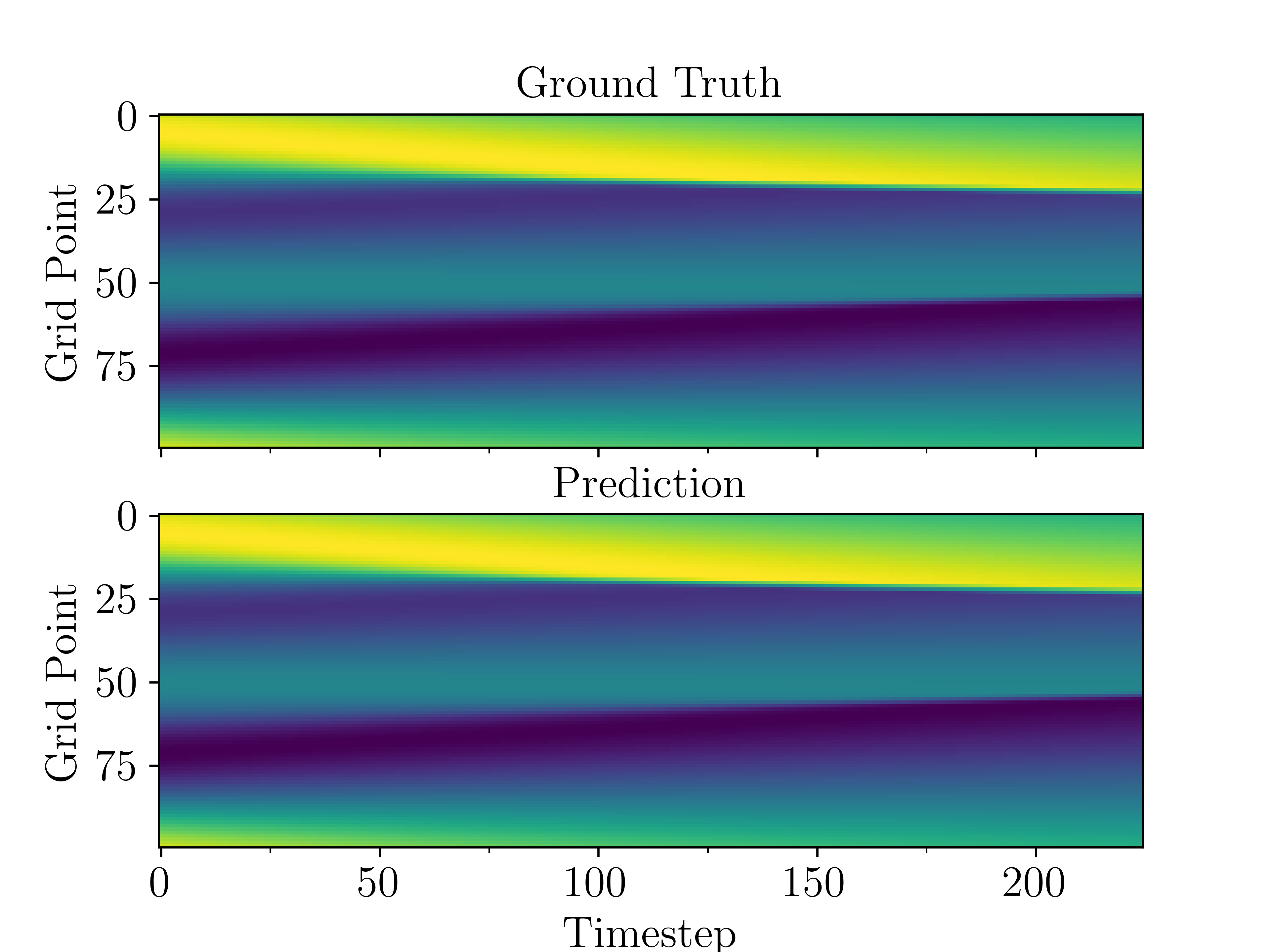}
\caption{Experiment \textbf{E1}, MSMP-PDE prediction}
\label{fig:figureE1}

\end{figure}

\begin{figure}[h]

\centering
\includegraphics[width=.5\textwidth]{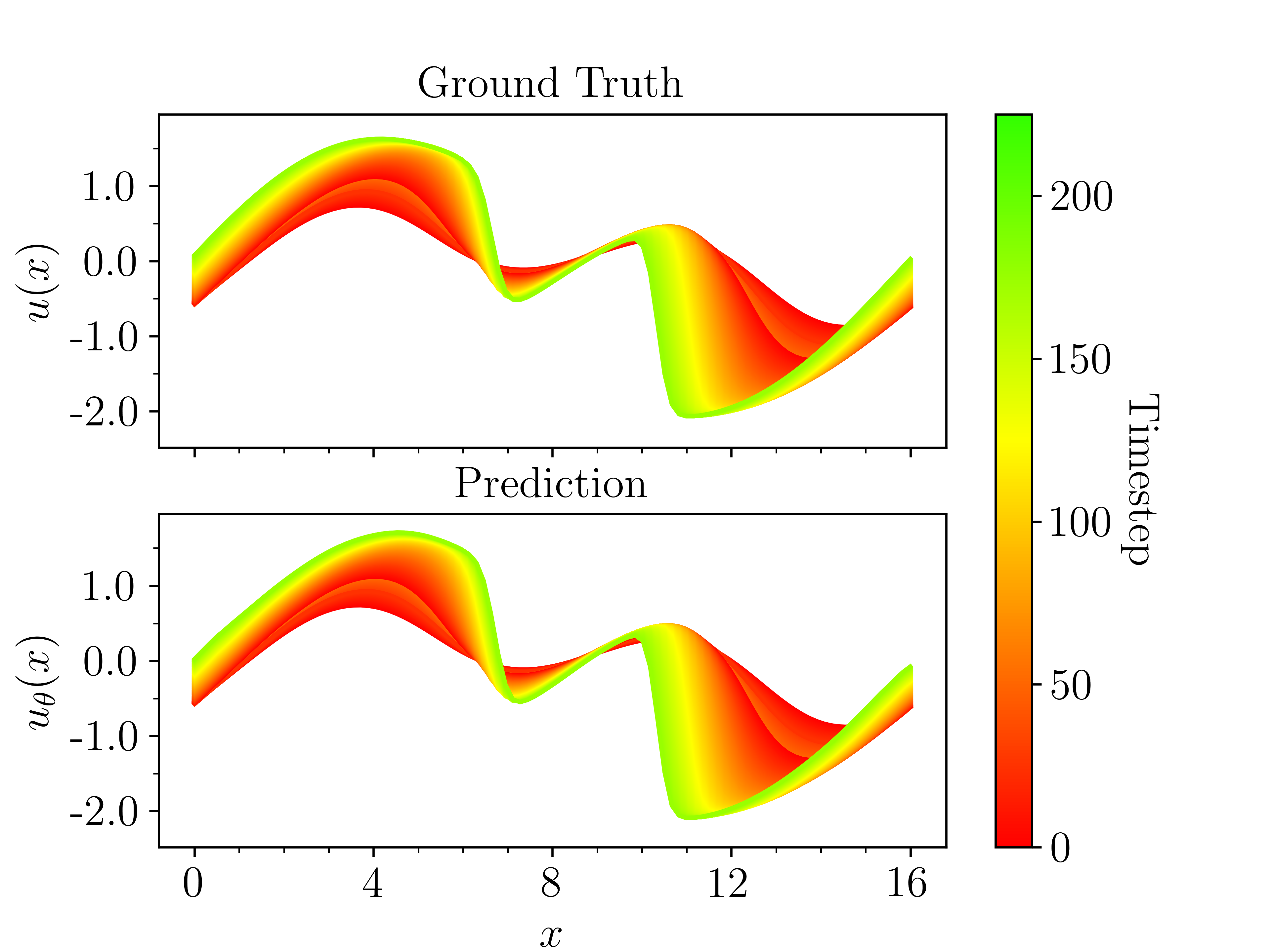}\hfill
\includegraphics[width=.5\textwidth]{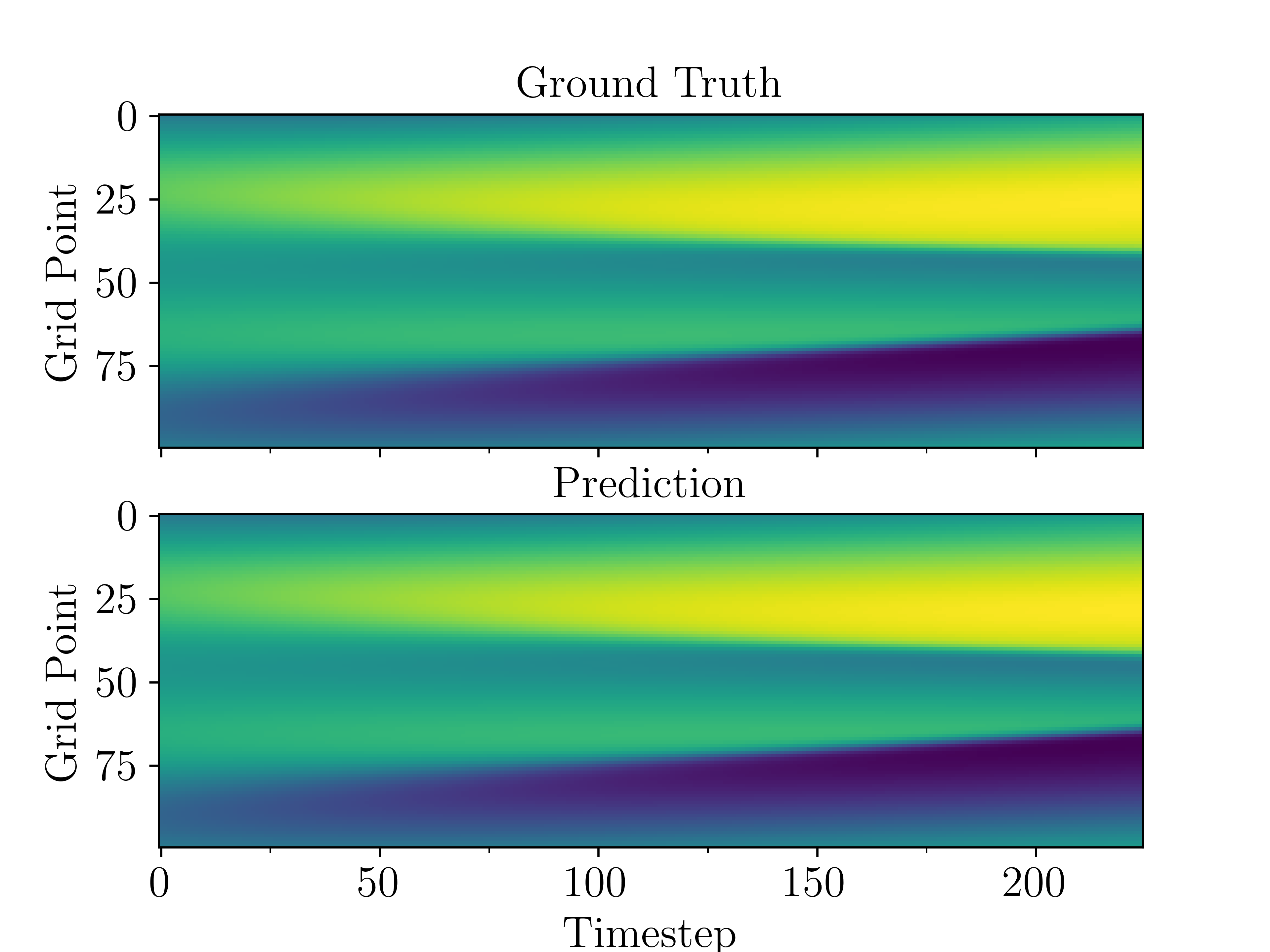}
\caption{Experiment \textbf{E2}, MSMP-PDE prediction}
\label{fig:figureE2}
\end{figure}

\begin{figure}[h]

\centering
\includegraphics[width=\textwidth]{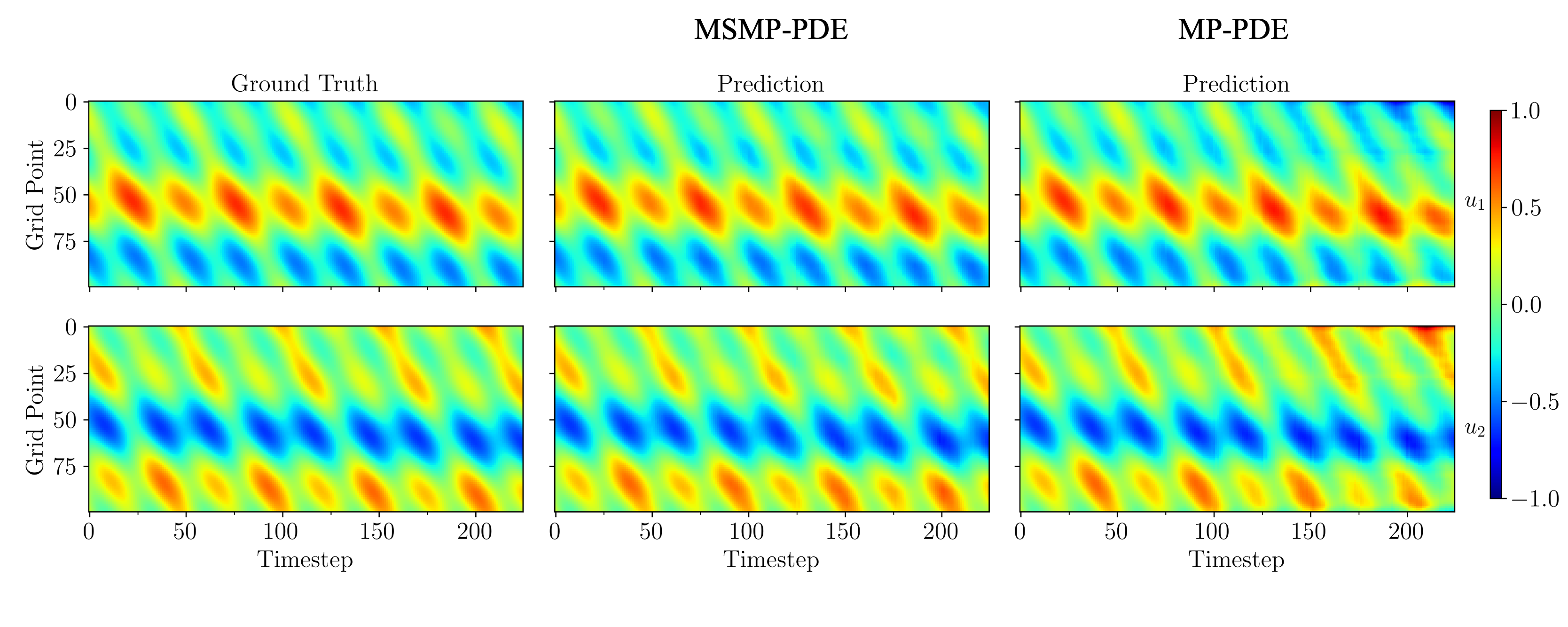}\hfill

\caption{Experiment \textbf{MS-wave} with equation parameters $a = 0.21$ and $b = 9.42$. MSMP-PDE shows a relative $L^2$ error of $9.6 \%$ while MP-PDE shows a relative $L^2$ error of $26.7\%$.}

\label{fig:figurecompRP}

\end{figure}

\begin{figure}[h]

\centering
\includegraphics[width=\textwidth]{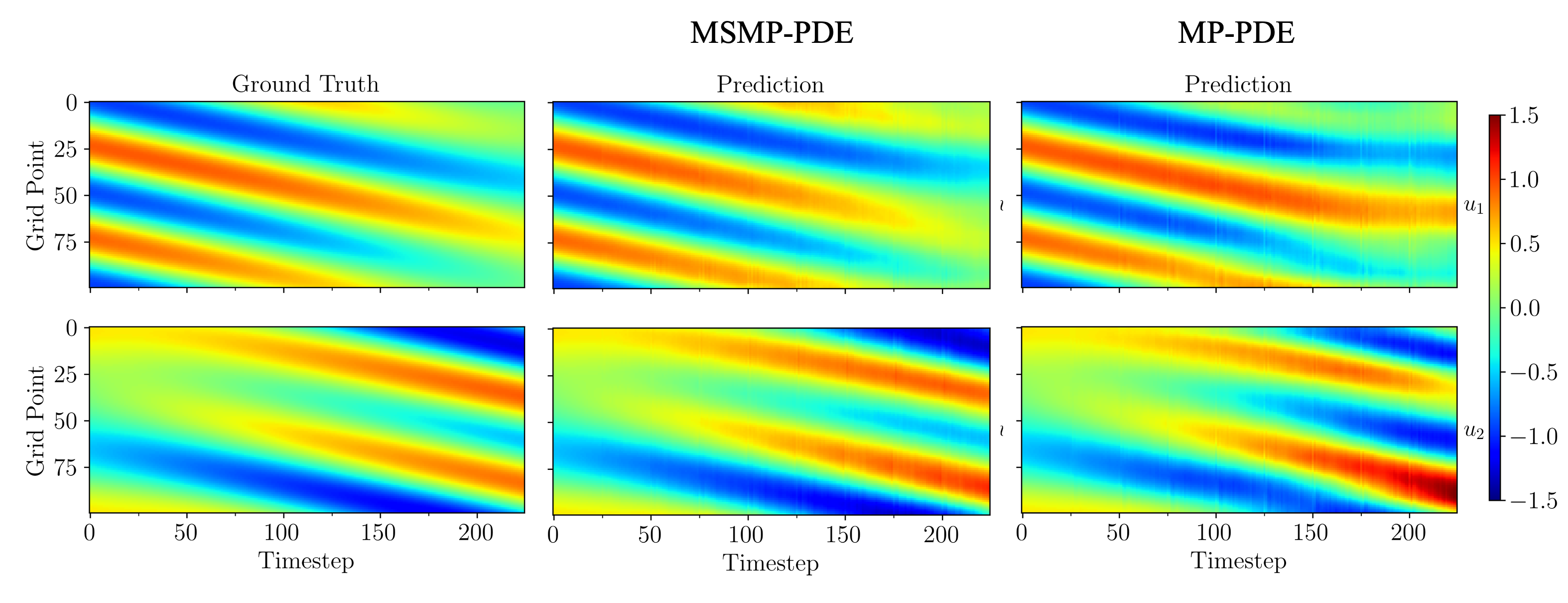}\hfill

\caption{Experiment \textbf{MS-wave} with equation parameters $a = 0.87$ and $b = 1.31$. MSMP-PDE shows a relative $L^2$ error of $14.05 \%$ while MP-PDE shows a relative $L^2$ error of $44.8\%$.}

\label{fig:figurecompRP2}

\end{figure}

\end{document}